\documentclass[runningheads]{llncs}

\usepackage{graphicx}
\usepackage{booktabs}

\usepackage[accsupp]{axessibility}  

\usepackage{hyperref}

\usepackage{orcidlink}

\usepackage{multirow}
\usepackage{booktabs}
\usepackage{makecell}
\usepackage{hhline}
\usepackage{subcaption}
\usepackage{xcolor}
\usepackage{multicol}
\usepackage{comment}

\definecolor{darkgreen}{rgb}{0,0.5,0}

\begin{document}

\sloppy

\title{EEG2Vision: A Multimodal EEG-Based Framework for 2D Visual Reconstruction in Cognitive Neuroscience}

\titlerunning{EEG2Vision}

\author{Emanuele Balloni\inst{1}\orcidlink{0000-0002-9510-5758} \and
Emanuele Frontoni\inst{2}\orcidlink{0000-0002-8893-9244} \and
Chiara Matti\inst{3} \and
Marina Paolanti\inst{2}\orcidlink{0000-0002-5523-7174} \and
Roberto Pierdicca\inst{1}\orcidlink{0000-0002-9160-834X} \and
Emiliano Santarnecchi\inst{4}\orcidlink{0000-0002-6533-7427}}

\authorrunning{E.~Balloni et al.}

\institute{Department of Civil Engineering Building and Architecture (DICEA), Università Politecnica delle Marche, Via Brecce Bianche 12, Ancona, 60131, Italy \and
Department of Political Sciences, Communication and International Relations, University of Macerata, Via Don Minzoni 22A, Macerata, 62100, Italy \and
Horizon Intelligence Labs \and
Harvard Medical School (HMS), Precision Neuromodulation Program \& Network Control Laboratory, Gordon Center for Medical Imaging, Department of Radiology, Massachusetts General Hospital, Horizon Intelligence Labs, Boston, MA, USA}

\maketitle

\begin{abstract}

Reconstructing visual stimuli from non-invasive electroencephalography (EEG) remains challenging due to its low spatial resolution and high noise, particularly under realistic low-density electrode configurations. To address this, we present EEG2Vision, a modular, end-to-end EEG-to-image framework that systematically evaluates reconstruction performance across different EEG resolutions (128, 64, 32, and 24 channels) and enhances visual quality through a prompt-guided post-reconstruction boosting mechanism. Starting from EEG-conditioned diffusion reconstruction, the boosting stage uses a multimodal large language model to extract semantic descriptions and leverages image-to-image diffusion to refine geometry and perceptual coherence while preserving EEG-grounded structure. Our experiments show that semantic decoding accuracy degrades significantly with channel reduction (e.g., 50-way Top-1 Acc from 89\% to 38\%),
while reconstruction quality slight decreases (e.g., FID from 76.77 to 80.51). The proposed boosting consistently improves perceptual metrics across all configurations, achieving up to 9.71\% IS gains in low-channel settings. A user study confirms the clear perceptual preference for boosted reconstructions. The proposed approach significantly boosts the feasibility of real-time brain-2-image applications using low-resolution EEG devices, potentially unlocking this type of applications outside laboratory settings.

\keywords{Artificial Intelligence \and Generative AI \and Neuroscience \and Computer Vision \and EEG \and EEG-to-image}
\end{abstract}

\section{Introduction}
\label{sec:introduction}
\textcolor{black}{Understanding how visual information is represented within the human brain is a fundamental objective of neuroscience \cite{doerig2025high,ozkirli2025computational,ferrante2025towards}. Decoding these representations from neural activity not only yields important insights into the computational mechanisms underlying visual perception \cite{zheng2020evoked}, but also opens concrete avenues for application. In brain-computer interface (BCI) research, the ability to reconstruct visual content from brain signals could enable new communication pathways for individuals with paralysis or severe motor impairments, as well as facilitate interactions that transcend language barriers \cite{guenther2024image}. Beyond clinical contexts, such capabilities motivate the longer-term possibility of prompting generative AI systems not through text or manual input, but directly through a user's visual intent.}
A key challenge lies in determining the extent to which internal perceptual representations can be inferred from non-invasive neural recordings~\cite{lu2021multi,oota2023deep}. 
Although functional magnetic resonance imaging (fMRI) has traditionally been the dominant technology in this field, thanks to its high spatial resolution~\cite{ren2021reconstructing,huo2024neuropictor,deng2024study}, practical constraints, such as low temporal resolution, high cost and lack of portability, limit its potential for real-time or widespread application. Electroencephalography (EEG) offers a portable and affordable alternative, capable of capturing neural dynamics at millisecond timescales. However, reconstructing visual stimuli from EEG is highly challenging due to the low signal-to-noise ratio, as well as the fact that scalp signals reflect the superposition of multiple synchronous cortical sources~\cite{spampinato2017deep,palazzo2020decoding,singh2024learning,lopez2025guess}.
To address these challenges, the field has rapidly evolved from simple classification to complex generation. Early approaches relied on compact CNN architectures originally developed for motor imagery \cite{lawhern2018eegnet,palazzo2020decoding,mishra2022eeg,spampinato2017deep}, demonstrating that spatiotemporal representations could be learned from non-invasive signals. These encoders were subsequently integrated into generative frameworks, such as GANs and VAEs, to synthesize images at the category level \cite{kavasidis2017brain2image,li2020semi,khare2022neurovision,jiao2019decoding}. Methods like ThoughtViz \cite{tirupattur2018thoughtviz} and attention-based variants \cite{mishra2023neurogan} further improved robustness through recurrent modeling and latent regularization \cite{singh2023eeg2image}.
Recent advances have been driven by the adoption of Latent Diffusion Models (LDMs), which provide expressive latent spaces capable of capturing fine-grained visual structures. Pioneering works, such as DreamDiffusion \cite{bai2023dreamdiffusion}, aligned EEG embeddings with CLIP space to condition Stable Diffusion \cite{patil2022stable}, significantly outperforming adversarial methods. Subsequent models, including MindDiffuser \cite{lu2023minddiffuser}, NeuroDM \cite{qian2024neurodm} and NeuroImagen \cite{lan2023seeing}, enhanced fidelity using transformer-based encoders. Currently, the state-of-the-art is represented by lightweight approaches like Guess What I Think (GWIT) \cite{lopez2025guess}, which injects EEG-derived spatial residuals into a frozen diffusion backbone via ControlNet.

Despite several promising developments, significant gaps remain in the current literature. In particular, EEG-to-image reconstruction methods are almost exclusively evaluated using high-density EEG (64-128 channels)~\cite{singh2024learning,lopez2025guess}. Although these configurations maximize spatial information, they are not feasible for real-world applications, where 16-32 channels are more common. 
Furthermore, no systematic analysis has been conducted on reconstruction performance across different electrode densities, nor on the contribution of specific electrodes or cortical regions to decoding accuracy. Few studies have explored low-density configurations~\cite{guenther2024image,li2024visual}, with evaluations usually restricted to fixed montages. This leaves the impact of progressive channel reduction largely unexplored.
Additionally, reconstructions often contain distortions, artefacts, or ambiguous details. This is because the EEG conditioning signal is weak and noisy~\cite{radford2021learning,rombach2022high,lopez2025guess}. Recent works primarily focus on improving conditioning mechanisms or EEG encoders, rather than incorporating dedicated post-reconstruction enhancement strategies. Diffusion models have strong generative priors, but without structured refinement, there is little control over how errors in EEG conditioning propagate into visual artifacts.

To address these limitations, we present EEG2Vision, a modular and integrated framework for reconstructing visual stimuli from brain activity. EEG2Vision is designed to provide a unified and rigorous assessment of the effect of EEG sensor density (from 128 to 24 channels) on reconstruction quality, while introducing a refinement mechanism that improves perceptual fidelity within realistic non-invasive neural constraints. The framework is organized as a single end-to-end pipeline, comprising interconnected modules for prior generation, neural embedding extraction, diffusion-based image generation and post-reconstruction boosting. It is specifically designed to evaluate the limits and create corrective strategies for EEG-driven visual reconstruction.

The main contributions are as follows: (i) The introduction of EEG2Vision, a unified EEG-to-image framework that combines a systematic electrode-density evaluation with a post-reconstruction enhancement stage; (ii) A rigorous analysis on the effectiveness of different EEG configuration on both classification and reconstruction stages; (iii) Ablation studies to quantify the contribution of individual electrodes and cortical regions, and the relevance of the Classifier-free Guidance (CFG) scale in the diffusion-based reconstruction process; (iv) The introduction of an effective, lightweight image-boosting mechanism that combines Multimodal Large Language Model (MLLM)-based semantic extraction with image-to-image diffusion to enhance geometry, texture and perceptual coherence while preserving EEG-grounded semantics. 

\section{Materials and methods}
\label{sec:materials}

\begin{figure}[htb!]
    \centering
    \includegraphics[width=\linewidth]{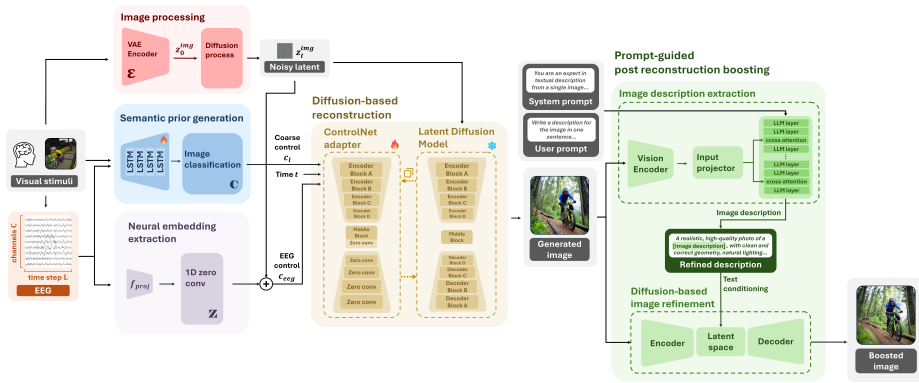}
    \caption{Overview of the EEG2Vision framework. The visual stimulus is encoded into a noisy latent via the \textit{Image processing} module and also fed into the \textit{Semantic prior generation} module, which utilizes both visual and EEG signals to derive coarse class-level controls. Simultaneously, EEG signals undergo \textit{Neural embedding extraction} to produce fine-grained spatiotemporal features. These inputs condition the \textit{Diffusion-based reconstruction} module through ControlNet to synthesize the generated image. Finally, the \textit{Prompt-guided post-reconstruction boosting} module refines the output via text-guided image-to-image diffusion.}
    \label{fig:framework_over}
\end{figure}

EEG2Vision is a modular, end-to-end system that combines semantic prior generation, neural embedding extraction, diffusion-based image generation, and multimodal refinement into a single, comprehensive process. It leverages state-of-the-art methods in EEG-to-image, LDM and MLLM, with the goal of generating coherent images from EEG signals and evaluating them thoroughly. The framework (in Fig.~\ref{fig:framework_over}) comprises five interconnected components: (i) an image processing step to encode the image and perform forward diffusion to obtain a latent representation; (ii) an auxiliary semantic module that generates coarse, class-level descriptions of the EEG signal to stabilize conditioning; (iii) a neural embedding module that produces compact temporal features from EEG signals; (iv) a diffusion-based reconstruction module that uses a ControlNet-modulated LDM; (v) a post-reconstruction enhancement stage that leverages MLLMs and image-to-image diffusion refinement to improve perceptual accuracy. These modules enable EEG2Vision to operate across different electrode densities while maintaining consistent decoding and reconstruction processes.

To examine the role of EEG spatial resolution, we trained and evaluated the framework under four different EEG spatial sampling densities: 128, 64, 32, and 24 channels. Reduced configurations were obtained by subsampling electrodes in a way that preserved symmetric bilateral coverage of frontal, temporal, parietal, and occipital regions, ensuring that lower-density montages remained neurophysiologically meaningful according to clinical and research standards (e.g., 10-20 system).
The following subsections detail each component of EEG2Vision.


\subsection{Image processing}

Starting from the ground truth (GT) image, we first apply an image processing step to obtain its latent representation in the diffusion model’s feature space.

Formally, let $\{y, x\}$ be a paired sample from the dataset, where
$y \in \mathbb{R}^{C \times L}$ denotes an EEG signal with $C$ channels and $L$ temporal samples, and
$x \in \mathbb{R}^{H_x \times W_x \times 3}$ is the corresponding visual stimulus. Following GWIT, the image $x$ is processed using a pretrained LDM and encoded into the latent space via a VAE encoder:
\begin{equation}
z^{img}_0 \sim \mathcal{E}_{\mathrm{VAE}}(x),
\quad
z^{img}_0 \in \mathbb{R}^{H_z \times W_z \times D}.
\end{equation}
Then, a forward diffusion process progressively adds Gaussian noise:
\begin{equation}
z^{img}_t = \sqrt{\alpha_t} z^{img}_0 + \sqrt{1 - \alpha_t}\,\epsilon,
\quad \epsilon \sim \mathcal{N}(0, I),
\end{equation}
where $t \in [0, T]$ denotes the diffusion timestep.
The objective is to reconstruct $z^{img}_0$ from $z^{img}_t$ conditioned on EEG-derived controls.

\subsection{Semantic prior generation}
\label{subsec:semantic_gen}

We introduce a coarse semantic control $c_l$ to stabilize conditioning and provide high-level guidance, following GWIT.
This control is obtained using a pretrained and frozen EEG image decoder (LSTM from \cite{singh2024learning} in particular), which predicts the stimulus class label from the EEG signal.
In particular, given the EEG input $y$, the decoder outputs a class label $\hat{l}$, which is converted into a textual caption of the form:
$c_l = $ ``Image of  $\hat{l}$''. The caption is encoded using the frozen text encoder of the LDM and used as an additional conditioning signal.
The EEG image decoder has been retrained independently for each EEG density configuration (128, 64, 32, and 24 channels) to maintain reliable semantic priors under reduced spatial resolution and perform coherent evaluation.

\subsection{Neural embedding extraction}

To align with the diffusion model’s latent space, raw EEG signals are transformed into compact neural embeddings. Specifically, EEG signals are projected directly into the latent image space rather than into a global embedding.
A projection network $f_{\mathrm{proj}}$ maps the EEG signal to a spatial latent tensor:
\begin{equation}
z^{eeg} = f_{\mathrm{proj}}(y),
\quad
f_{\mathrm{proj}} : \mathbb{R}^{C \times L} \rightarrow \mathbb{R}^{H_z \times W_z \times D}.
\end{equation}
The projection network consists of stacked 1D convolutional layers operating along the temporal dimension, followed by reshaping to match the spatial dimensions of the image latent.
To prevent harmful interference with the pretrained diffusion backbone during early training, the EEG latent is passed through a zero-initialized $1 \times 1$ convolution $Z(\cdot)$, as in ControlNet:
\begin{equation}
c_{\mathrm{eeg}} = z^{img}_t + Z(z^{eeg}).
\end{equation}
This EEG control tensor serves as the primary conditioning input to the ControlNet adapter.

\subsection{Diffusion-based reconstruction}
\label{subsec:diffusion_rec}

EEG conditioning is injected using a ControlNet adapter.
Let the LDM UNet consist of an encoder $E_{\theta}$ and a decoder $D_{\theta}$.
The ControlNet adapter is defined as a trainable copy of the UNet encoder, denoted $E_{\theta'}$.
Given the EEG control $c_{\mathrm{eeg}}$, coarse control $c_l$, and timestep $t$, the ControlNet processes the conditioning as:
\begin{equation}
\mathrm{ControlNet}(c_{\mathrm{eeg}}, c_l, t)
=
E_{\theta'}(c_{\mathrm{eeg}}, c_l, t)
=
E_{\theta'}(z^{img}_t + Z(f_{\mathrm{proj}}(y)), c_l, t).
\end{equation}
The ControlNet adapter produces feature residuals that modulate the frozen UNet backbone during denoising.
The training objective follows the standard diffusion loss, following the original approach:
\begin{equation}
\mathcal{L}
=
\mathbb{E}_{z^{img}_t, z^{eeg}, c_l, t, \epsilon}
\left[
\left\|
\epsilon -
\epsilon_{\theta}
\left(
z^{img}_t,
z^{eeg},
c_l,
t
\right)
\right\|_2^2
\right],
\end{equation}
where $\epsilon_{\theta}$ denotes the noise prediction of the UNet.
During training, the UNet backbone remains frozen, and only $E_{\theta'}$ and $f_{\mathrm{proj}}$ are optimized.
Separate ControlNet models have been trained for each of the four EEG density configuration.

\subsection{Prompt-guided post-reconstruction boosting}
\label{sec:post_boosting}

Images generated from the EEG-conditioned LDM can exhibit blurred textures, missing fine-grained details, or structural artifacts, especially under reduced EEG channel configurations.
These limitations arise from the inherently low signal-to-noise ratio of EEG and the weak spatial constraints provided during the denoising process. To address this issues, we introduce a prompt-guided post-reconstruction boosting stage that enhances perceptual quality while preserving EEG-consistent semantics.
This stage is applied after the EEG-conditioned generation and, thus, does not alter the training or inference procedure performed in the previous steps, making it suitable as a model-agnostic approach.

\paragraph{Image description extraction}
Given an EEG-reconstructed image $\hat{x}$, we first extract an explicit semantic description using a pretrained MLLM (LLaMA 3.2 Vision \cite{grattafiori2024llama} in our case):
\begin{equation}
p = \mathcal{M}_{\text{MLLM}}(\hat{x}),
\end{equation}
where $p$ is a concise, single-sentence textual description of the visual content.
The MLLM is prompted with a fixed system and user instruction that enforces objective, image-grounded descriptions and prevents hallucinated content or stylistic embellishments. Tab. \ref{tab:prompt_mllm} describes the prompts in detail.
\begin{table}[htb]
\centering
\caption{Prompt template used for the MLLM to generate concise image descriptions.}
\label{tab:prompt_mllm}
\begin{tabular}{|p{0.95\linewidth}|}
\hline
\textbf{System Prompt}: "You are an expert in textual description 
from a single image. Given an image, you will provide a concise and 
accurate description of the content, without saying \lq the image shows' 
or \lq the image depicts' at the start."

\textbf{User Prompt}: "Write a description for this image in one 
sentence. You should answer with the prompt only. Do not insert the first part where you say \lq the image shows' or \lq the image depicts' in the answer."\\
\hline
\end{tabular}
\end{table}
This step converts implicit semantic information embedded in $\hat{x}$ into an explicit textual representation.

\paragraph{Diffusion-based image refinement}
The generated description $d$ is then integrated into a structured refinement prompt $p_{\text{ref}}$, designed to enforce constraints on geometry, textures, lighting, and visual quality, ensuring that the refined image both preserves the original EEG-driven structure and eliminates visual artifacts.
This prompt explicitly encodes visual quality constraints while grounding semantic content in the EEG-reconstructed image through $d$. Refer to Tab. \ref{tab:prompt_dm} for the full prompt.
\begin{table}[htb]
\centering
\caption{Prompt template used for the image-to-image diffusion model, conditioned on the image description.}
\label{tab:prompt_dm}
\begin{tabular}{|p{0.95\linewidth}|}
\hline
$p_{\text{ref}} = $ "A realistic, high-quality photo of a $[d]$, 
with clean and correct geometry, natural lighting, 
consistent textures, and accurate proportions. 
No visual glitches, no distorted shapes, no rendering artifacts. 
The object appears physically plausible and professionally photographed, 
with all structures logically and realistically aligned."\\
\hline
\end{tabular}
\end{table}

The reconstructed image $\hat{x}$ and the composed prompt $p_{\text{ref}}$ are subsequently used to condition an image-to-image LDM (Stable Diffusion 3 Medium \cite{patil2022stable}):
\begin{equation}
\tilde{x}
=
\mathcal{D}_{\text{img2img}}
\big(
\hat{x}
\;\big|\;
\mathrm{TextEnc}(p_{\text{ref}})
\big),
\end{equation}
where $\hat{x}$ provides structural initialization and $p_{\text{ref}}$ supplies high-level semantic and perceptual guidance via the text encoder.
The noise strength of the image-to-image process is carefully controlled to limit deviations from $\hat{x}$, ensuring that refinement focuses on correcting artifacts and enhancing visual fidelity rather than altering semantic content.

\subsection{Evaluation protocol}

We evaluate the EEG-to-image synthesis pipeline across three stages: semantic decoding, diffusion-based reconstruction, and post-generation boosting, with the goal of isolating the effect of EEG channel resolution on categorical grounding and generative quality. 
To validate the EEG-to-class decoder (Sec. \ref{subsec:semantic_gen}), we report $N$-way Top-$k$ Accuracy \cite{lan2023seeing,singh2024learning,lopez2025guess}. Using a pretrained ImageNet classifier, this metric measures the ability to map EEG latents to the correct ground-truth visual category. In particular, we report 50-way Top-1 and Top-5 Accuracy.
To evaluate ControlNet-guided diffusion (Sec. \ref{subsec:diffusion_rec}) and the boosting stage (Sec. \ref{sec:post_boosting}), we use Inception Score (IS) \cite{salimans2016improved} and Fr\'{e}chet Inception Distance (FID)~\cite{heusel2017gans} to assess distributional similarity and image quality, and Learned Perceptual Image Patch Similarity (LPIPS)~\cite{zhang2018unreasonable} to measure perceptual similarity between reconstructions and ground-truth stimuli. Additionally, CLIP Cosine Similarity (CLIP-Sim)~\cite{radford2021learning} is leveraged to evaluate the preservation of high-level semantic features.
For the prompt-guided boosting stage, we report only IS, FID, and LPIPS. Classification accuracy is omitted because boosting operates on fixed reconstructions without altering EEG decoding. CLIP-Sim is also excluded to avoid bias, as prompt-driven refinement could artificially increase text–image alignment scores without improving EEG-grounded faithfulness.

\section{Results and Discussion}
\label{sec:results}

\subsection{Experimental settings}
\label{sec:exp_settings}

\paragraph{Dataset}
We employ the EEGCVPR40 dataset for our experiments~\cite{spampinato2017deep,palazzo2020decoding}. This dataset is one of the most widely used benchmark for EEG-to-image reconstruction \cite{singh2024learning,lopez2025guess,bai2023dreamdiffusion}. It contains EEG recordings collected from 6 participants while they viewed a total of 2,000 images drawn from 40 distinct object categories of the ImageNet database \cite{deng2009imagenet}. Each category includes 50 images, presented sequentially for 0.5 seconds per trial at 1 kHz. After every block of 50 images, a 10-second pause was introduced to reduce fatigue and allow for reset. EEG activity was recorded with a 128-channel system (ActiCAP 128ch). The stimulus set encompassed a diverse range of visual categories, including animals, vehicles, and everyday objects, ensuring broad semantic coverage. EEG data pre-processing was performed following \cite{lopez2025guess}. Moreover, data have been split as the original implementation \cite{spampinato2017deep}, with training, validation and test sets corresponding to 80\%, 10\% and 10\% of the dataset, respectively.

\paragraph{Implementation details} Experiments were conducted on a system with an NVIDIA H100 GPU (80~GB VRAM), AMD EPYC 9V84 96-core CPU, and 320~GB RAM, running Ubuntu 22.04 LTS. The environment used was Python 3.9 and PyTorch 2.1.0 with CUDA 12.1. Llama 3.2 Vision was employed for MLLM and Stable Diffusion 3 Medium (\textit{img2img}) for diffusion-based refinement. The EEG-to-image model (GWIT) followed the original hyperparameters: LSTM-based decoder trained for 8192 epochs, batch size 256, learning rate \(3 \times 10^{-4}\). For CFG, we compare $\gamma=4$ and $\gamma=7.5$, where $\gamma=4$ is the default CFG used in GWIT, and $\gamma=7.5$ reflects common practice in diffusion-based image generation \cite{patil2022stable}. The ControlNet adapter was trained with learning rate \(1 \times 10^{-5}\), Adam optimizer, batch size 32, for 100 epochs, monitoring validation to prevent overfitting. For evaluation, we generated 4 samples per test-set EEG trial for each channel configuration (128, 64, 32, and 24 channels). All metrics were computed on the entire EEGCVPR40 test set.

\subsection{Diffusion-based reconstruction results}
\label{sec:eeg-to-image_res}

Quantitative results, reported in Tab.~\ref{tab:quant_results_diff}, reveal a general trend of performance degradation as the number of electrodes is reduced. This relationship is modulated significantly by the CFG and exhibits non-linear characteristics.
\begin{table}[htb]
\centering
\caption{Quantitative results of different channel configurations and CFGs across multiple performance metrics.}
\label{tab:quant_results_diff}
\resizebox{.8\textwidth}{!}{%
\setlength{\tabcolsep}{6pt}
\begin{tabular}{@{}llccccc ccc@{}}
\toprule
\multirow{2}{*}[-3pt]{\textbf{Channels}} & \multirow{2}{*}[-3pt]{\textbf{CFG}} & \multicolumn{2}{c}{\textbf{50-way Accuracy}} & \multirow{2}{*}[-3pt]{\textbf{IS $\uparrow$}} & \multirow{2}{*}[-3pt]{\textbf{FID $\downarrow$}} & \multirow{2}{*}[-3pt]{\textbf{LPIPS $\downarrow$}} & \multirow{2}{*}[-3pt]{\textbf{CLIP-Sim $\uparrow$}} \\
\cmidrule(lr){3-4}
& & \textbf{Top-1 $\uparrow$} & \textbf{Top-5 $\uparrow$} & & & & \\
\midrule
\multirow{2}{*}{\textbf{128}} & \textbf{4} & 0.876 & 0.928 & 33.93 & 79.14 & 0.770 & 0.723 \\
                     & \textbf{7.5} & \textbf{0.890} & \textbf{0.935} & \textbf{34.82} & \textbf{76.77} & \textbf{0.770} & \textbf{0.733} \\
\midrule
\multirow{2}{*}{\textbf{64}}  & \textbf{4} & 0.800 & 0.869 & 33.45 & 83.05 & 0.773 & 0.700 \\
                     & \textbf{7.5} & \textbf{0.823} & \textbf{0.877} & \textbf{34.11} & \textbf{78.27} & \textbf{0.777} & \textbf{0.714} \\
\midrule
\multirow{2}{*}{\textbf{32}}  & \textbf{4} & 0.375 & 0.458 & 33.26 & 86.00 & 0.785 & 0.620 \\
                     & \textbf{7.5} & \textbf{0.386} & \textbf{0.468} & \textbf{33.28} & \textbf{80.55} & \textbf{0.791} & \textbf{0.629} \\
\midrule
\multirow{2}{*}{\textbf{24}}  & \textbf{4} & 0.371 & 0.446 & 33.70 & 84.09 & 0.787 & 0.617 \\
                     & \textbf{7.5} & \textbf{0.380} & \textbf{0.450} & \textbf{34.24} & \textbf{80.51} & \textbf{0.790} & \textbf{0.625} \\
\bottomrule
\end{tabular}}
\end{table}
The most pronounced effect of reducing channel count is seen in the image classification accuracy. The 50-way Top-1 accuracy
drops significantly from $89\%$ with 128 channels to $38\%$ with 24 channels when using a CFG of $\gamma=7.5$. This shows that the spatial detail captured by high-density arrays is crucial for robust semantic decoding. The sharpest decline occurs between the 64-channel and 32-channel configurations, highlighting a critical threshold below which the neural signal becomes too weak for fine-grained classification. A degradation can also be seen, in part, in the image reconstruction quality. Metrics that assess the perceptual and distributional similarity between generated and GT images perform worse with decreasing channel count.
Nevertheless, the performance loss is not as impactful as the one in image classification. FID, for instance, increases from 76.77 to 80.51 between the 128-channel and 24-channel configurations, a change of less than 5 points. Similarly, CLIP-Sim, which measures high-level semantic alignment, shows only a slight decline, from a value of 0.733 for the 128 channels configuration to 0.625 for the 24 channels one. Furthermore, IS remained remarkably stable across all channel reductions. This suggests that the diversity and basic recognizability of the generated images are preserved by the frozen LDM backbone, even when the conditioning EEG signal is severely compromised. Nevertheles, the performance loss is noticeable, especially in low-channel settings.
A key factor in mitigating the impact of channel reduction is the CFG. The consistent superiority of the higher guidance scale ($\gamma=7.5$) over the ($\gamma=4$) value across all configurations indicates that the coarse semantic prior from the text prompt becomes increasingly vital. As the fine-grained neural features weaken, amplifying the influence of this prior helps anchor the generation process, ensuring the output remains semantically grounded and of high quality.
Another interesting result is the performance of the 24-channel configuration, which performs on par with or even marginally surpasses the 32-channel setup on certain metrics, like IS and FID. This non-monotonic behavior can be attributed to the electrode subsampling strategy. The 24-channel montage was deliberately designed to preserve symmetric coverage of neurophysiologically relevant regions, whereas the 32-channel selection may have included a less optimal or more redundant set of electrodes. This implies that, for low-channel systems, the strategic placement of electrodes to cover key brain networks may be just as important as the absolute number of channels, and a well-designed low-density montage can potentially outperform a poorly designed higher-density one.
\begin{figure}[htb]
    \centering
    \includegraphics[width=.7\linewidth]{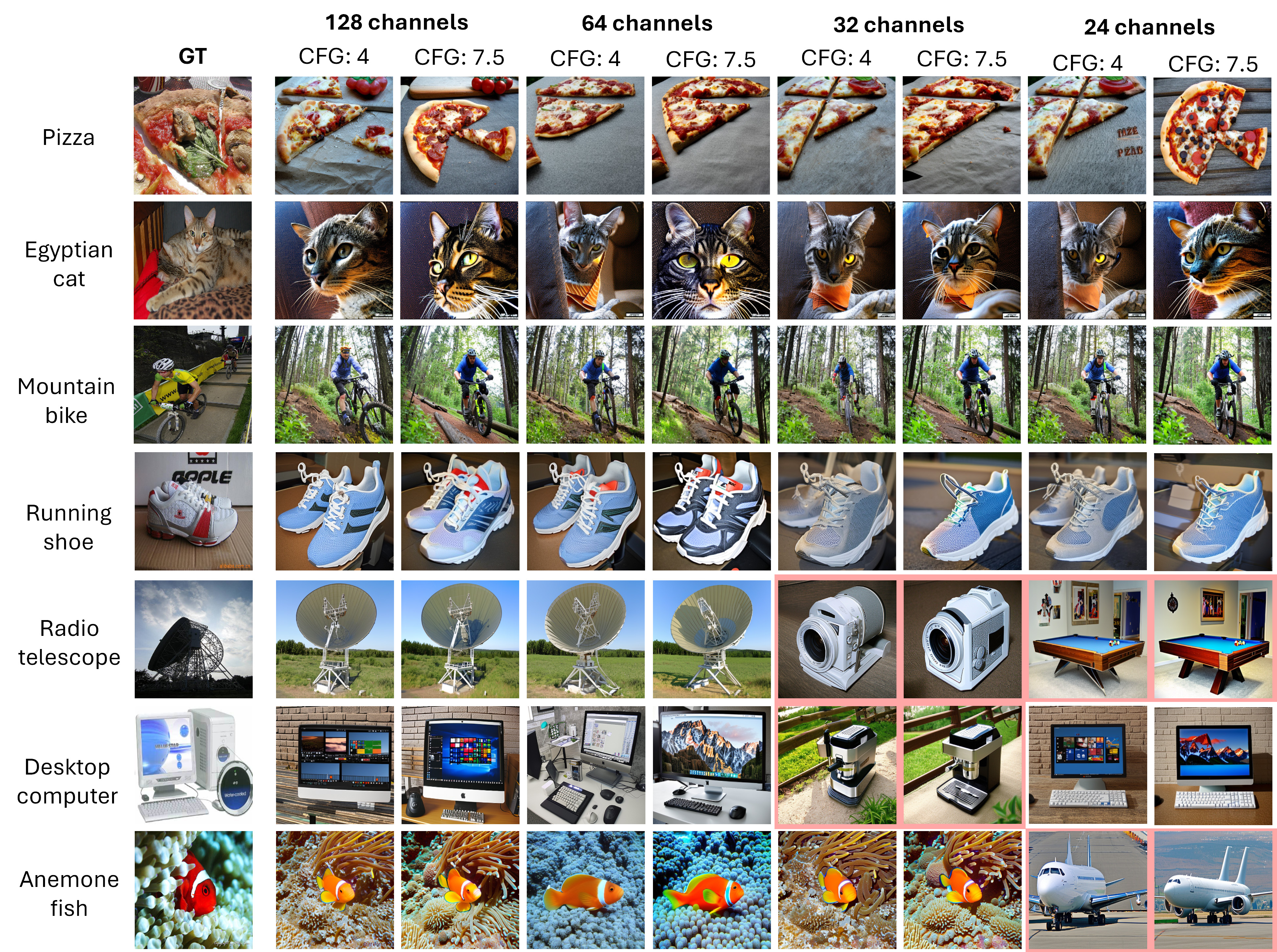}
    \caption{Qualitative results of the evaluation for different channel resolutions and CFG configurations. Some failure cases are also shown for low accuracy classes (highlighted in red).}
    \label{fig:qual_results_diff}
\end{figure}
Qualitative results in Fig. \ref{fig:qual_results_diff} further support these observations. Reconstructions obtained with 128 channels exhibit sharp structures, accurate geometry, and well-defined textures. As channel density decreases, images become progressively noisier and less detailed, with more frequent misclassifications and structural distortions at 32 and 24 channels.
Nevertheless, the pipeline retains significant functionality, even with low-channel systems for image reconstruction.

\subsection{Prompt-guided post-reconstruction boosting results}
\label{sec:image_boosting_res}
We evaluated the performance of our proposed reconstruction image boosting by comparing the quality of raw EEG reconstructions, in its best CFG scale configuration $\gamma=7.5$, against the boosted images produced after MLLM-guided diffusion refinement.
\begin{table}[htb!]
\centering
\caption{Quantitative evaluation of raw and boosted reconstructions across different electrode configurations. Percentage gains indicate improvement of the boosted method over the original.}
\label{tab:quant_results}
\resizebox{\textwidth}{!}{%
\begin{tabular}{lccc@{\hspace{5pt}}ccc@{\hspace{5pt}}ccc@{\hspace{5pt}}ccc}
\toprule
 & \multicolumn{3}{c}{\textbf{128 Channels}} 
 & \multicolumn{3}{c}{\textbf{64 Channels}} 
 & \multicolumn{3}{c}{\textbf{32 Channels}} 
 & \multicolumn{3}{c}{\textbf{24 Channels}} \\
\cmidrule(lr){2-4} \cmidrule(lr){5-7} \cmidrule(lr){8-10} \cmidrule(lr){11-13}
\textbf{Metric} 
& \textbf{Raw} & \textbf{Boosted} & \textbf{Gain}
& \textbf{Raw} & \textbf{Boosted} & \textbf{Gain}
& \textbf{Raw} & \textbf{Boosted} & \textbf{Gain}
& \textbf{Raw} & \textbf{Boosted} & \textbf{Gain} \\
\midrule
\textbf{IS $\uparrow$}
& 34.82 & \textbf{37.23} & \textcolor{darkgreen}{+6.69\%}
& 34.11 & \textbf{37.37} & \textcolor{darkgreen}{+9.12\%}
& 33.28 & \textbf{36.51} & \textcolor{darkgreen}{+9.71\%}
& 34.24 & \textbf{37.16} & \textcolor{darkgreen}{+8.53\%} \\
\textbf{FID $\downarrow$}
& 76.77 & \textbf{77.06} & \textcolor{darkgreen}{+0.38\%}
& 78.27 & \textbf{77.75} & \textcolor{darkgreen}{+0.67\%}
& 80.55 & \textbf{79.26} & \textcolor{darkgreen}{+1.60\%}
& 80.51 & \textbf{79.77} & \textcolor{darkgreen}{+0.92\%} \\
\textbf{LPIPS $\downarrow$}
& 0.770 & \textbf{0.769} & \textcolor{darkgreen}{+0.13\%}
& 0.777 & \textbf{0.773} & \textcolor{darkgreen}{+0.52\%}
& 0.791 & \textbf{0.787} & \textcolor{darkgreen}{+0.51\%}
& 0.790 & \textbf{0.785} & \textcolor{darkgreen}{+0.63\%} \\
\bottomrule
\end{tabular}}
\end{table}
Tab.~\ref{tab:quant_results} presents the quantitative results across all electrode configurations using the established image quality metrics.
As shown, our framework consistently improved all metrics across all channel counts, demonstrating its effectiveness in enhancing visual fidelity. The IS saw the most substantial gains, with improvements ranging from +6.69\% to +9.71\%, indicating that the boosted images are more semantically meaningful and diverse. FID and LPIPS scores also decreased consistently across all configurations.
In par with the original reconstruction, the 24-channel configuration generally showed better boosted reconstruction results compared to the 32-channel setup across most metrics. Furthermore, the relative gains provided by our boosting framework are generally more substantial in lower-channel configurations. For instance, the improvement in FID is greatest (+1.60\%) for the 32-channel setup, which also has the worst raw FID score. This trend is expected, as reconstructions from high-density EEG (e.g., 128 channels) already contain rich information and exhibit fewer artifacts, leaving less room for drastic improvement. In contrast, low-channel setups suffer from significant information loss and artifacts, providing a greater opportunity for our multimodal refinement stage to correct errors and enhance detail. Nevertheless, the results confirm that our framework can be beneficially applied to any channel configuration to yield images with higher visual quality and semantic fidelity.
\begin{figure}[htb]
    \centering
    \includegraphics[width=.7\linewidth]{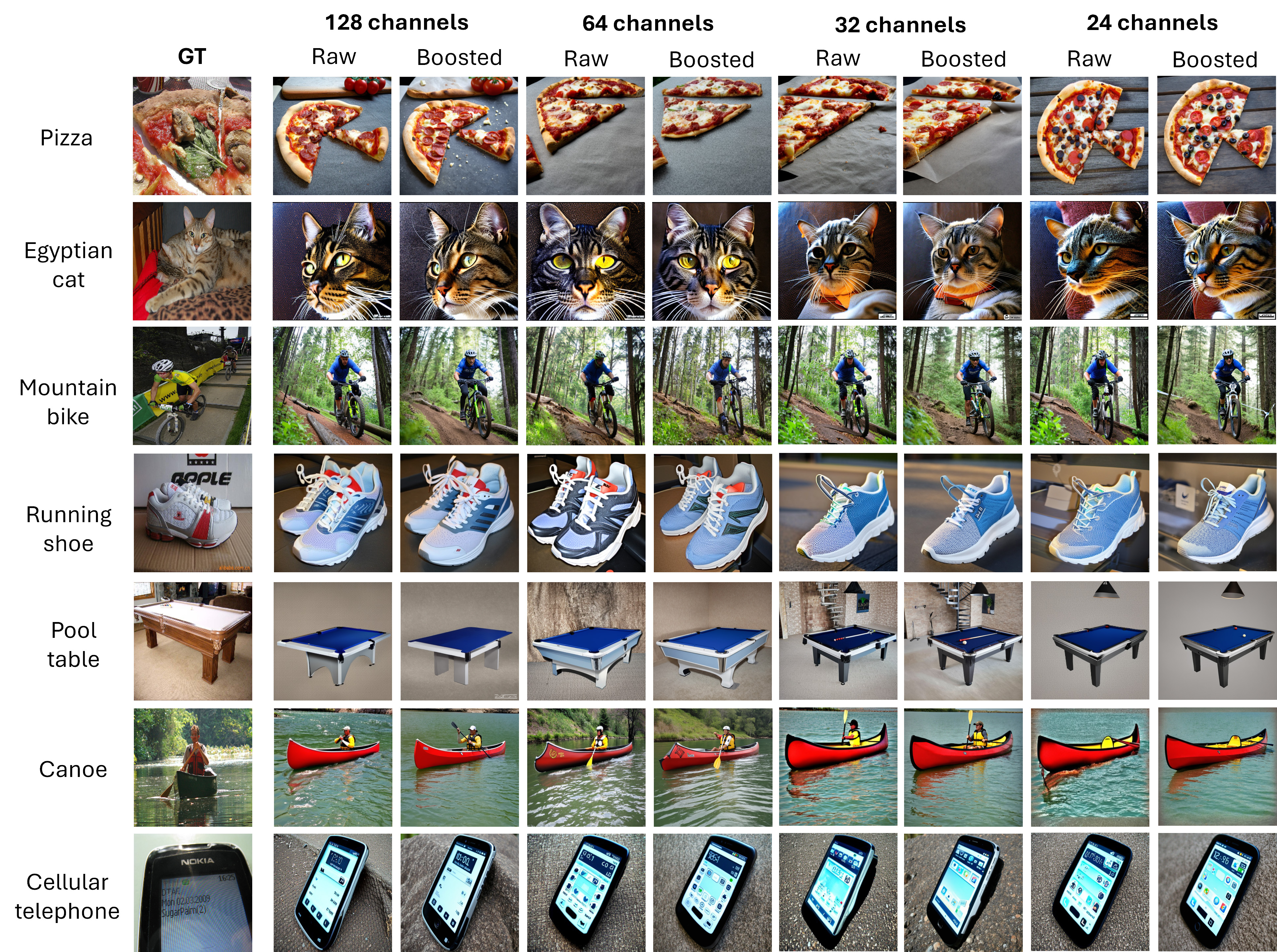}
    \caption{Qualitative comparison of reconstructions across electrode configurations. Columns from left to right show: GT reference image, raw generated reconstruction and our refined output (boosted).}
    \label{fig:qual_comp}
\end{figure}
The qualitative improvements achieved by our boosting framework are visually demonstrated in Fig.~\ref{fig:qual_comp}, showcasing example reconstructions across all channel configurations alongside their corresponding GT images. A visual inspection of these results reveals consistent patterns that underscore the effectiveness of our approach and align with the quantitative findings.

These results confirm that the MLLM-guided diffusion process successfully integrates high-level semantic priors with the structural information from the initial EEG decoding, producing images that are not only more realistic but also more faithful to the original visual stimuli that generated the neural responses.

\subsection{Analysis of brain regions}
\label{sec:electrode_importance}

\begin{figure}[htb]
    \centering
    \begin{minipage}[b]{.58\textwidth}
        \centering
        
        \begin{subfigure}[b]{0.48\textwidth}
            \centering
            \includegraphics[width=.9\textwidth]{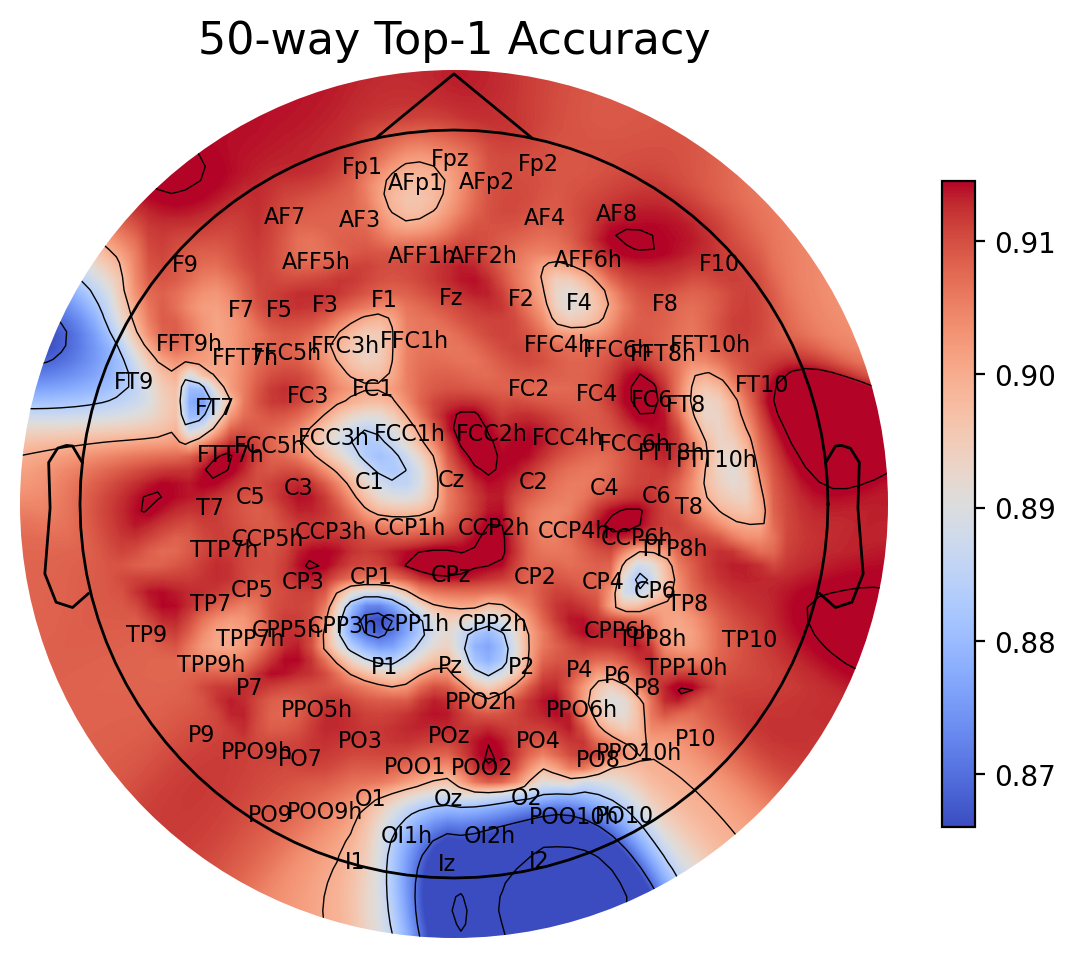}

            \vspace{0.4em}
            
            \includegraphics[width=\textwidth]{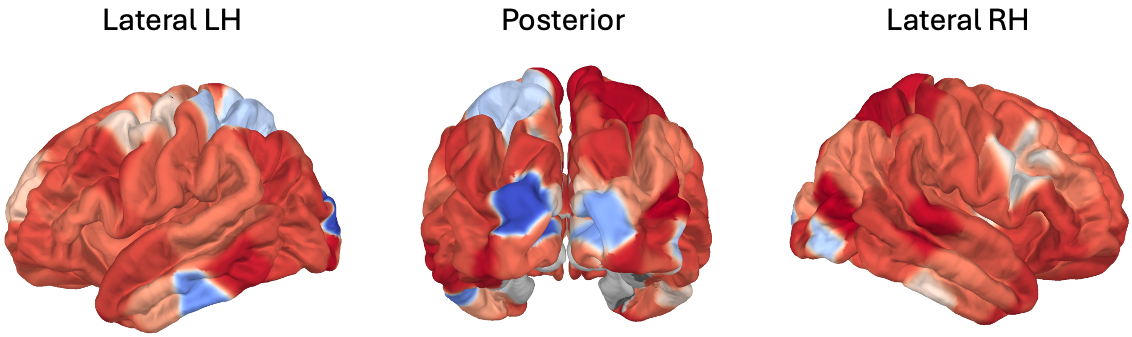}
            \caption{Top-1 Acc}
            \label{fig:topo_top1}
        \end{subfigure}
        \hfill
        \begin{subfigure}[b]{0.48\textwidth}
            \centering
            \includegraphics[width=.9\textwidth]{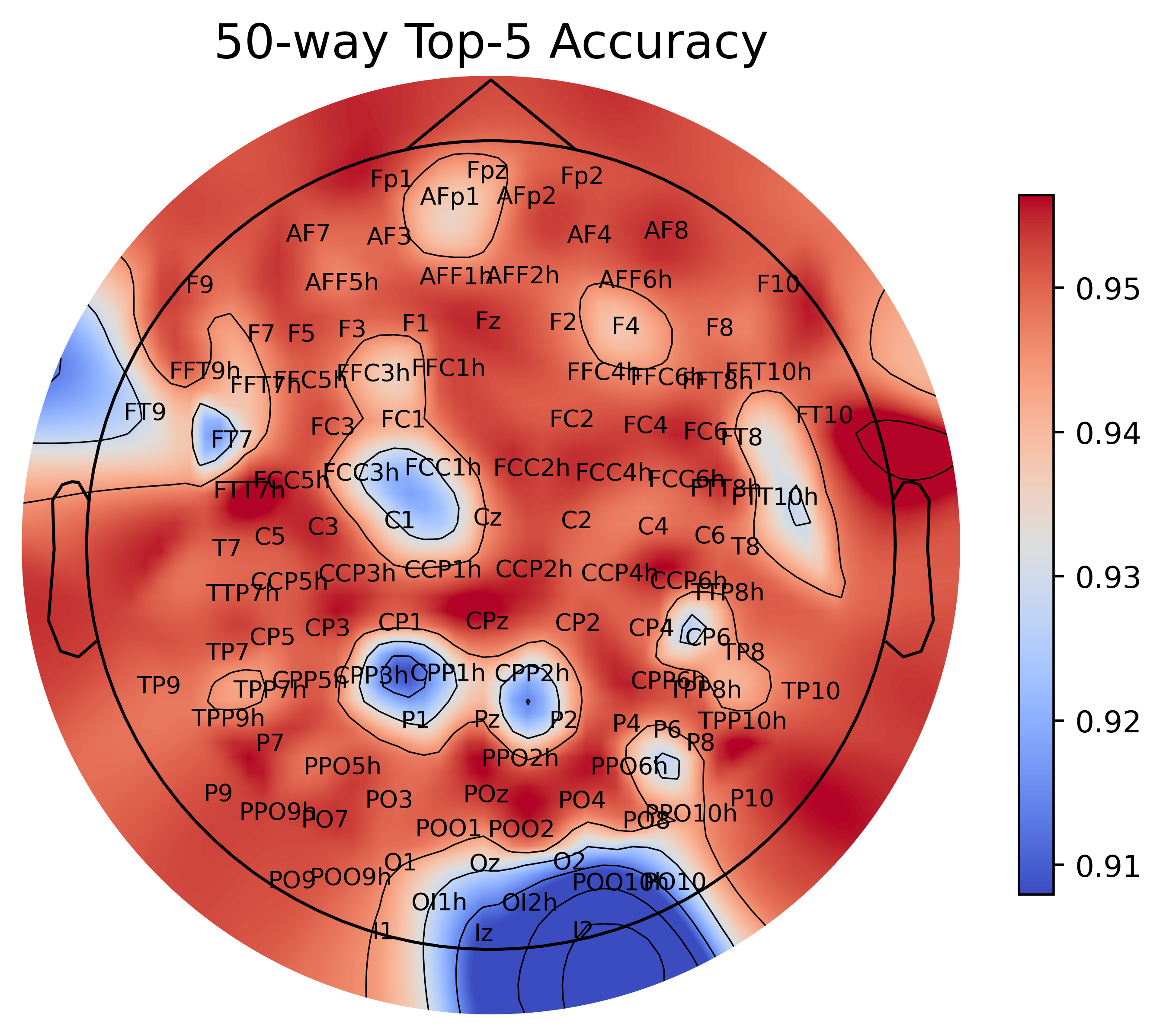}

            \vspace{0.4em}
            
            \includegraphics[width=\textwidth]{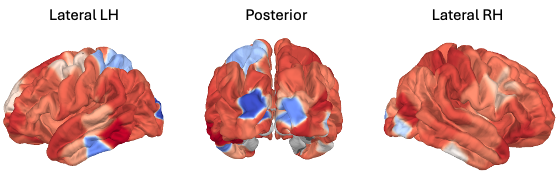}
            \caption{Top-5 Acc}
            \label{fig:topo_top5}
        \end{subfigure}

        \caption{Spatial distribution of decoding accuracy per electrode.}
        \label{fig:topoplots}
    \end{minipage}
    \hfill
    \begin{minipage}[b]{0.4\textwidth}
        \centering
        \begin{subfigure}[b]{.65\textwidth}
            \centering
            \includegraphics[width=\textwidth]{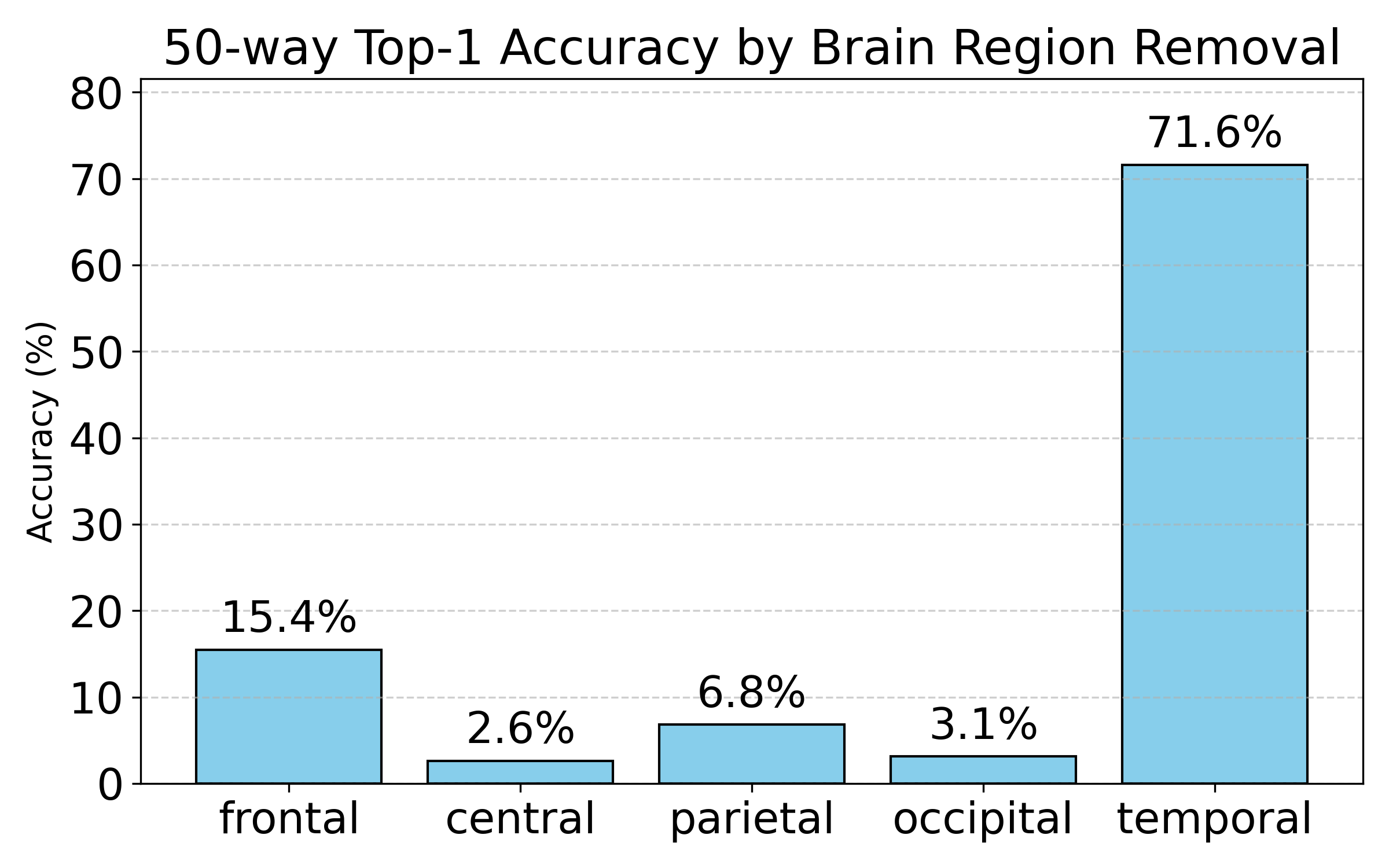}
            \caption{Top-1 Acc}
            \label{fig:bar_top1}
        \end{subfigure}
        \\
        \begin{subfigure}[b]{.65\textwidth}
            \centering
            \includegraphics[width=\textwidth]{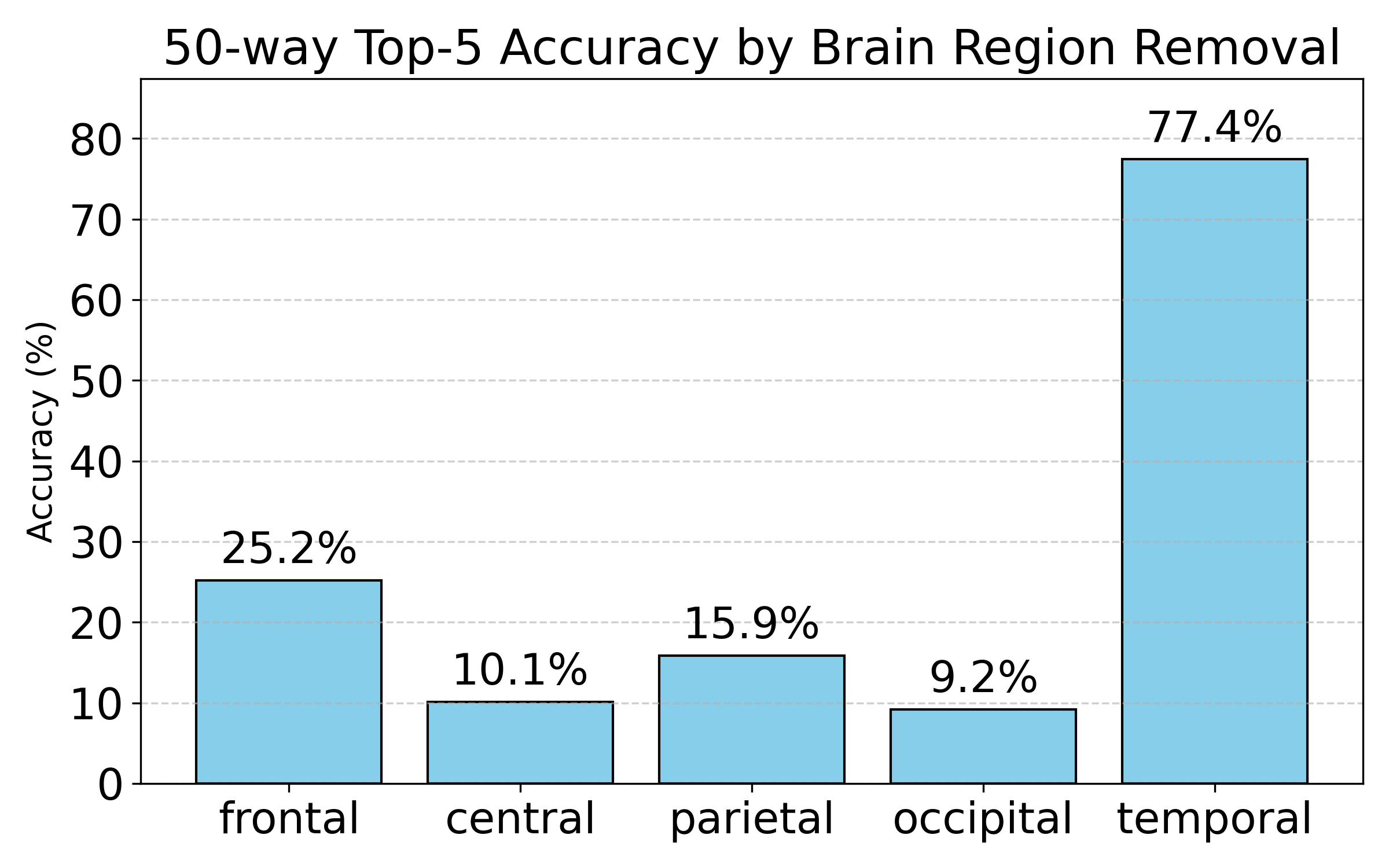}
            \caption{Top-5 Acc}
            \label{fig:bar_top5}
        \end{subfigure}
        \caption{Impact on decoding accuracy upon region removal.}
        \label{fig:region_barplots}
    \end{minipage}
\end{figure}

To evaluate the contribution of specific neural sources to the image generation process, we conducted a detailed ablation study on the 128-channel setting, following \cite{guo2025neuro}. This analysis was designed to quantify the impact of individual electrodes and broader brain regions on the model's classification accuracy, providing insights into the spatial distribution of visually relevant information within the EEG signal. The analysis consisted of two complementary parts: per-electrode analysis and region-based analysis.

To assess the contribution of individual electrodes, we computed 50-way Top-1 and Top-5 accuracy for each channel in isolation. For a given electrode, the activity from that single channel was set to zero and given as input to the EEG image decoder, together with all other channels, unaltered. This procedure was repeated for every electrode, generating a spatial accuracy map across the scalp. Results are visualized in Fig.~\ref{fig:topoplots}. The Top-1 accuracy topoplot reveals a well-defined spatial pattern, with the most impactful performance loss localized towards the occipital and central regions. A similar pattern, albeit with generally higher accuracy values, is observed in the Top-5 accuracy topoplot. This distribution aligns with previous research on the topic \cite{grill2001lateral,li2024visual,song2023decoding} and is anatomically coherent with the location of occipital electrodes over visual cortex. In addition, considering that each image in the dataset is presented for only 0.5 s, the decoding process primarily relies on neural activity occurring within the first stages of visual perception associated to the initial feedforward sweep of visual information. Consistently, early visual evoked potentials typically originate in occipital electrodes and emerge the first 100-200 ms following stimulus onset \cite{dirusso2002}.



To evaluate the importance of broader functional brain networks, we also performed a region ablation study. We grouped electrodes into five standard brain networks: frontal, central, parietal, occipital, and temporal. The analysis involved systematically removing all electrodes within one region and evaluating the model's performance using the remaining channels. This measured the performance drop attributable to the loss of information from that specific region. The results, presented in Fig.~\ref{fig:region_barplots}, reveal that the occipital region is the most critical for visual classification. Its removal resulted in the most severe performance degradation, with Top-1 accuracy dropping to just $3.1\%$ and Top-5 accuracy to $9.2\%$. This large loss is consistent with the fundamental role of the primary visual cortex in in extracting stimulus features during the first stages of perception. Similarly, the removal of the parietal region, known for attentional selection \cite{bisley_attention_2010}, also caused a substantial decrease in performance (Top-1: $6.8\%$, Top-5: $15.9\%$). The removal of the central region also greatly affected accuracy results, with Top-1 dropping to $2.6\%$ and Top-5 to $10.1\%$. This result is particularly noteworthy given that central electrodes are often sensitive to attentional modulation and early feedback processes. Event-related potential studies show that visual categorization and attentional influences can already emerge around 150 ms after stimulus onset, indicating that top-down processes start interacting with sensory representations very early in the processing cascade \cite{thorpe_speed_1996}. These mechanisms may help stabilize task-relevant representations and therefore support EEG-based decoding. 
In contrast, the removal of the frontal and temporal regions resulted in relatively minor declines in accuracy. In particular, the temporal region removal reduces accuracy by just $17.4\%$ and $16.1\%$ for Top-1 and Top-5, respectively. 
This suggests that, while these regions may contribute to broader cognitive processes, their information content is either less critical for the specific task of image category classification or is redundantly encoded in other areas. Another possible factor is the concrete nature of the stimulus categories used: a recent study \cite{fares2025} reports that frontal and temporal networks are preferentially engaged during the processing of abstract semantic content, whereas concrete visual stimuli primarily recruit occipital and parietal regions. This interpretation, however, should be verified with stimulus sets that include more abstract categories.

\section{User Study}
\label{sec:user_study}

To complement the metrics and qualitative analysis of the prompt-guided post-reconstruction boosting, we conducted a comprehensive perceptual evaluation involving 20 human subjects. The study was designed to validate whether the improvements measured by quantitative metrics would be perceptible and meaningful to external observers. We ran a two-alternative forced-choice experiment. For each trial, participants viewed the raw reconstruction and the corresponding boosted output in a side-by-side configuration with randomized left-right placement and blinded category assignment. They were instructed to select the preferred image and to rate their confidence level on a 5-point Likert scale. Trials covered all EEG montages used in our evaluation (24, 32, 64, 128 channels), enabling stratified analyses by electrode density under otherwise identical viewing conditions. This design follows established pairwise preference protocols for perceptual comparisons \cite{prashnani2018pieapp,chen2023gap,miao2008quantitative}, while the confidence rating provides an interpretable weight for aggregating choices.
A total of 554 evaluable trials have been acquired. Ages ranged from 23 to 64 years, with a mean of approximately 38.6 years across respondents, and education skewed toward advanced degrees (Master’s and PhD), with some representation from Bachelor’s and professional degrees. We also asked participants about their experience with generative AI (on a 5-point Likert scale): mean GenAI familiarity was 4.0, mean usage of GenAI image-generation tools was 3.0, and mean confidence in spotting AI-generated images was 3.0
We evaluated the results with a boosted preference rate metric, which is defined as the fraction of trials in which the boosted image was chosen against the total. The confidence-weighted preference was also computed to reflect metacognitive certainty. This metric weights the preference rate for each choice based on the confidence score. For each montage, we report the unweighted boosted preference rate, the confidence-weighted preference rate, and the mean confidence. We tested overall boosted preference against a 50\% chance using a two-sided normal approximation to the binomial to concisely summarize perceptual advantage.
\begin{table}[htb!]
\centering
\caption{User study results showing preference for boosted images across different channel configurations. Confidence ratings are on a 5-point Likert scale.}
\label{tab:user_study}
\resizebox{.9\textwidth}{!}{%
\setlength{\tabcolsep}{6pt}
\begin{tabular}{lccc}
\toprule
\textbf{Channels} & \textbf{Boosted Preference Rate} & \textbf{Mean Confidence} & \textbf{Weighted Preference Rate} \\
\midrule
128 & 71.43\% & 3.7 & 74.17\% \\
64 & 82.35\% & 3.5 & 85.35\% \\
32 & 76.97\% & 3.6 & 79\% \\
24 & 82.73\% & 3.6 & 87.13\% \\
\hline
\textbf{Overall} & 78.34\% & 3.6 & 81.31\% \\
\bottomrule
\end{tabular}}
\end{table}

Results are reported in Tab. \ref{tab:user_study}. Across all trials and participants, boosted images were preferred in 78.34\% of comparisons, rising to 81.31\% when weighting by confidence. The mean confidence was 3.6. A binomial test against chance (50\%) indicates a highly significant deviation in favor of the boosted images, confirming a robust perceptual advantage. Stratifying by channel count shows consistent positive margins across 24, 32, 64, and 128 channels. While magnitudes vary by montage, the qualitative pattern remains stable. This indicates that boosting improves perceived quality in both low and high-density EEG.
Together, these findings provide evidence that the boosting stage improvements are meaningful to the observers.

\section{Conclusions and future works}
\label{sec:conclusion}

We presented EEG2Vision, a framework for evaluating EEG-conditioned image synthesis under realistic channel constraints, complemented by a lightweight post-reconstruction refinement to recover perceptual quality. Experiments from 128 to 24 channels show that, while semantic decoding degrades with reduced spatial sampling, perceptual and distributional metrics remain relatively stable when a frozen diffusion prior is guided by compact semantic structure and residual EEG modulation.
Electrode and regional ablations highlight that montage design is critical: preserving bilateral posterior coverage, particularly occipital sites with supportive central electrodes, is more effective than uniform downsampling. The prompt-guided boosting stage consistently improves geometry, textures, and artifact suppression across all configurations, with larger gains at lower channel counts.
\textcolor{black}{These results carry distinct implications depending on the application context. For neuroscientific investigations (e.g., examining how categorical boundaries in visual cortex map onto reconstructed image space, or comparing reconstructed representations across levels of the visual hierarchy) high-density acquisitions remain the appropriate choice, as semantic accuracy degrades substantially below 64 channels. At the same time, the viability of 24-channel configurations, when paired with the boosting stage, represents an important step toward out-of-laboratory deployment in BCI applications and, more generally, toward paradigms in which generative AI models are conditioned on neural signals rather than manual input. Nevertheless, both directions will require further investigation, especially at low channel densities, where inter-subject variability and signal reliability remain open challenges.}

Future developments will focus on improving robustness and adaptability. Subject-specific adaptation via few-shot calibration or lightweight adapters may reduce inter-subject variability, especially at low channel densities. Joint optimization of EEG encoding, diffusion conditioning, and refinement could limit semantic drift while preserving perceptual quality. Finally, incorporating auxiliary signals such as eye tracking or EOG may enhance attentional alignment in low-channel settings.

%
%
\bibliographystyle{splncs04}
\bibliography{main}
\end{document}